\documentclass[runningheads]{llncs}
\usepackage{graphicx}

\usepackage{tikz}
\usepackage{comment}
\usepackage{amsmath,amssymb} 
\usepackage{color}
\usepackage{makecell} 
\usepackage{enumitem}
\usepackage{extarrows}

\usepackage[pagebackref,breaklinks,colorlinks]{hyperref}

\usepackage{booktabs}
\usepackage{amsfonts}
\usepackage{bm}
\usepackage{arydshln}
\usepackage{multirow}

\usepackage{orcidlink}
\usepackage[misc]{ifsym} 

\usepackage[accsupp]{axessibility}  
\newcommand{\myparagraph}[1]{ \noindent \textbf{#1}}

\begin{document}
\pagestyle{headings}
\mainmatter
\def\ECCVSubNumber{820}  

\title{Semantic-Sparse Colorization Network for Deep Exemplar-based Colorization} 


\titlerunning{Semantic-Sparse Colorization Network for Deep Exemplar-based Colorization}
%
\author{Yunpeng Bai\inst{1} \and
Chao Dong \inst{2,3} \and
Zenghao Chai\inst{1}\orcidlink{0000-0003-3709-4947} \and
Andong Wang\inst{1} \and
Zhengzhuo Xu\inst{1} \and
Chun Yuan\inst{1,4}$^{(\textrm{\Letter})}$ }

\authorrunning{Bai et al.}
%
\institute{Tsinghua Shenzhen International Graduate School, China \and
Shenzhen Institutes of Advanced Technology, Chinese Academy of Sciences \and
Shanghai AI Laboratory, China \and
Peng Cheng National Laboratory, China \\
\email{\{byp20, wad20, xzz20\}@mails.tsinghua.edu.cn, chao.dong@siat.ac.cn, zenghaochai@gmail.com, yuanc@sz.tsinghua.edu.cn}}
\maketitle
\newcommand\blfootnote[1]{%
\begingroup
\renewcommand\thefootnote{}\footnote{#1}%
\addtocounter{footnote}{-1}%
\endgroup
}

\begin{abstract}
  Exemplar-based colorization approaches rely on reference image to provide plausible colors for target gray-scale image. The key and difficulty of exemplar-based colorization is to establish an accurate correspondence between these two images. Previous approaches have attempted to construct such a correspondence but are faced with two obstacles. First, using luminance channel for the calculation of correspondence is inaccurate. Second, the dense correspondence they built introduces wrong matching results and increases the computation burden. To address these two problems, we propose Semantic-Sparse Colorization Network (SSCN) to transfer both the global image style and detailed semantic-related colors to the gray-scale image in a coarse-to-fine manner. Our network can perfectly balance the global and local colors while alleviating the ambiguous matching problem. Experiments show that our method outperforms existing methods in both quantitative and qualitative evaluation and achieves state-of-the-art performance.
\keywords{image colorization, sparse attention, exemplar-based colorization}
\end{abstract}

\blfootnote{$^{\textrm{\Letter}}$ Corresponding author}

\section{Introduction}

Image colorization is a classic and appealing task that predicts the vivid colors from a gray-scale image. As there is no unique correct color for a given pixel, three classes of methods are proposed to constrain the output color space. The first one is called automatic colorization, such as \cite{DBLP:conf/iccv/ChengYS15,DBLP:conf/eccv/ZhangIE16}. These methods generally rely on the powerful convolutional networks and learn a direct mapping from a large-scale image dataset. The second class introduces additional human intervention, such as user-guided scribbles \cite{DBLP:journals/tog/ZhangZIGLYE17,DBLP:conf/cvpr/SangkloyLFYH17,DBLP:conf/mm/CiMWLL18} and text \cite{DBLP:conf/naacl/ManjunathaIBD18,DBLP:conf/eccv/BahngYCPWMC18}. They require users to provide reliable color/text labels for more dedicated colorization. While the third class, denoted as exemplar-based method \cite{DBLP:conf/mm/LuYPZW20,DBLP:journals/tog/HeCLSY18,DBLP:journals/cgf/XiaoHZQWHH20,DBLP:conf/cvpr/XuWFSZ20,DBLP:conf/cvpr/LeeKLKCC20,DBLP:journals/tip/BugeauTP14,DBLP:journals/tog/ChiaZGTCTL11,DBLP:conf/mm/GuptaCRNH12,DBLP:conf/mm/YinLZP21,DBLP:journals/tip/LiSLAC21}, is a trade-off between fully automatic and human intervention strategies. It adopts a reference image as guidance and generates a similar color-style image. These three kinds of methods have different applications and prior information, thus cannot be compared side-by-side. In this work, we study exemplar-based image colorization, due to its large flexibility and excellent performance.

\begin{figure}
  \centering
  \includegraphics[width=1\linewidth]{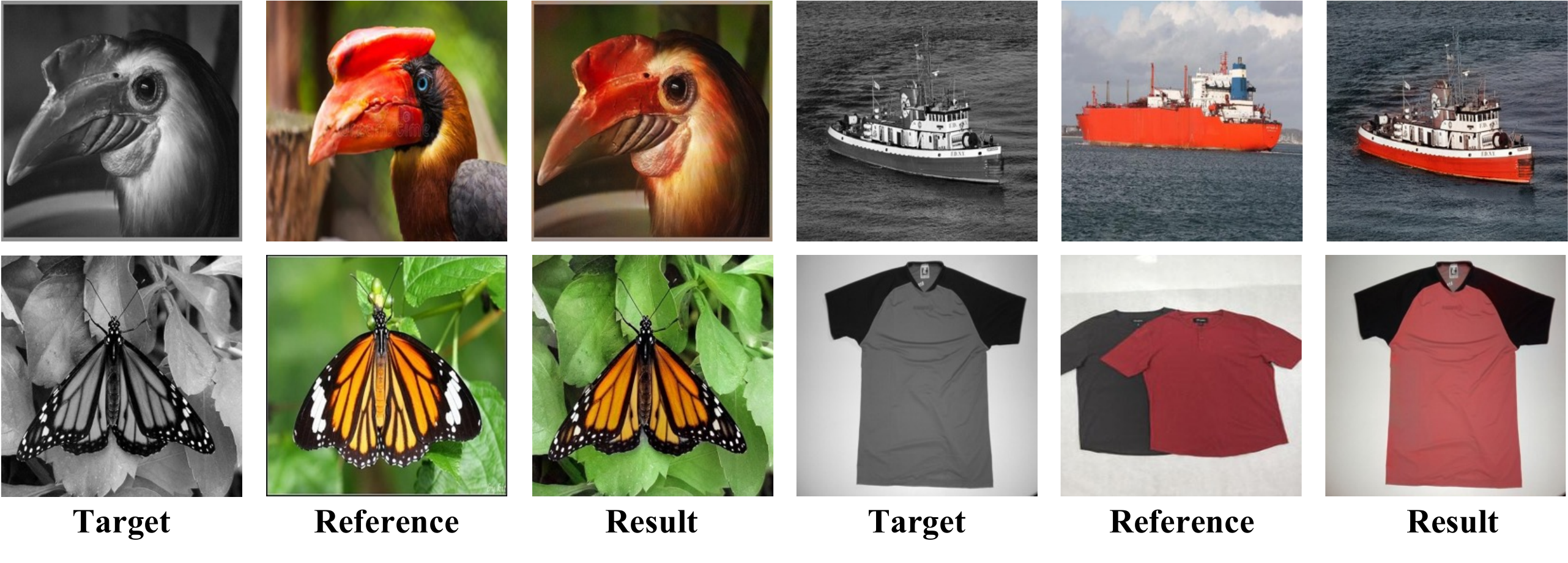}
  \caption{Overview colorization results of the proposed method. Our method can commendably build correspondence between the target and reference images and has the capability to generate a plausible colorization of gray-scale images.}
  \label{fig::overview}
\end{figure}

The difficulty of exemplar-based image colorization is to build an accurate correspondence between the gray-scale image and the color reference. 
Some works regard colorization as a style transfer problem \cite{DBLP:conf/cvpr/XuWFSZ20}, and usually transfer the global color tones. As a result, they lack detailed color matching between semantically similar objects/parts.
Other researchers \cite{DBLP:conf/cvpr/ZhangHLSYBC19,DBLP:conf/cvpr/LeeKLKCC20,DBLP:conf/mm/LuYPZW20,DBLP:conf/mm/YinLZP21,DBLP:journals/corr/abs-2201-04364} propose to construct a dense correspondence with a correlation matrix, whose elements characterize pairwise similarity between different image features. Although they have achieved considerable progress, they are still facing two obstacles. 
First, the correspondence is calculated using the luminance channel of the input image.
However, as gray-scale images do not contain enough semantic information as color images (a common knowledge in image classification \cite{DBLP:journals/tog/HeCLSY18}), the correspondence based on the luminance channel \cite{DBLP:conf/mm/LuYPZW20,DBLP:conf/mm/YinLZP21,DBLP:conf/cvpr/ZhangHLSYBC19,DBLP:journals/corr/abs-2201-04364} is inaccurate.
Second, the dense correspondence itself will also bring in unavoidable drawbacks. It not only introduces wrong matching results for semantically unrelated objects, but also increases the computation burden.

To address the above mentioned problems, we propose a new coarse-to-fine colorization framework -- Semantic-Sparse Colorization Network -- to transfer both the global image style and the detailed semantic-related colors to the gray-scale image. Specifically, in the coarse colorization stage, we adopt an image transfer network to obtain a preliminary colorized result. The color information of the reference image is encoded as a vector, which is then migrated to the gray-scale image by an AdaIN \cite{DBLP:conf/iccv/HuangB17} operation. In the fine colorization stage, we will first calculate the semantic correspondence between the coarse result and the reference image. Specially, only the semantic-significant parts and some background regions are reserved for calculation, leading to a sparse correlation matrix. Then the attention mechanism will be used to re-weight the reference image and help generate the final color result. The proposed method can perfectly balance the global and local colors while alleviating the ambiguous matching problem caused by dense correspondence. Extensive experiments have shown the superiority of our network towards other state-of-the-art methods. To facilitate numerical evaluation, we also propose a unified evaluation pipeline for all exemplar-based colorization methods. Our code will be publicly available for research purpose.

Our main contributions are summarized as follows:

\begin{itemize}

\item We propose to build a more accurate correspondence between a coarse-colorized result and the reference image. It not only minimizes the information gap between the gray-scale input and the color reference, but also achieves better performance on details.
\item We propose a sparse attention mechanism to make the model focus on the semantically significant regions in the reference image. It could produce more detailed results with lower computation cost.
\item We collect a new test dataset from ImageNet to solve the problem of fair comparison. We also design a new quantitative evaluation metric to evaluate exemplar-based colorization methods.

\end{itemize}

\section{Related Work}
Because image colorization plays an essential role in image processing tasks such as old photo restoration and image editing, this subject has been studied for a long time \cite{DBLP:conf/eccv/CharpiatHS08,DBLP:journals/tip/BugeauTP14,DBLP:journals/tog/QuWH06,DBLP:conf/rt/LuanWCLXS07,DBLP:conf/mm/HuangTCWW05}. 
Recently, many studies have used learning-based methods to solve this ill-posed problem. 
These approaches can be roughly grouped into three classes.

The first one is called automatic colorization, which directly maps gray-scale images to color images, such as
\cite{DBLP:conf/iccv/ChengYS15} and \cite{DBLP:conf/eccv/ZhangIE16}. 
They are the earliest methods to use convolutional networks to learn the mapping from a large-scale image dataset.
MemoPainter \cite{DBLP:conf/cvpr/YooBCLCC19} uses a memory network to “memorize” rare examples, which can avoid the interference of dominant color in the dataset and make the model perform well even without sufficient data. More recently, Transformer has also been applied to address this task \cite{DBLP:journals/corr/abs-2102-04432}. Some works \cite{DBLP:conf/pkdd/CaoZZY17,DBLP:conf/wacv/VitoriaRB20,DBLP:journals/corr/abs-2108-08826} take advantage of generative models to promote the diversity of results. For instance, \cite{DBLP:journals/corr/abs-2108-08826} leverages the rich and diverse color priors encapsulated in a pretrained StyleGAN \cite{DBLP:conf/cvpr/KarrasLA19} to recover vivid colors. The variational autoencoder (VAE) architecture has also been used in
 \cite{DBLP:conf/cvpr/DeshpandeLYCF17}. However, the colorization process of these methods are lack of controllability.

The second class introduces additional human intervention, such as user-guided scribbles and text. They require users to provide reliable color/text labels for more dedicated colorization. Traditional scribble-based colorization methods \cite{DBLP:journals/tog/LevinLW04,DBLP:journals/tog/XuLJHL09} usually propagate the local user hints to the whole image via an optimization approach, while learning-based methods
\cite{DBLP:journals/tog/ZhangZIGLYE17,DBLP:conf/cvpr/SangkloyLFYH17,DBLP:conf/mm/CiMWLL18} will combine color prior learned from large-scale image dataset with user's intervention for colorization. 
Recently, some researchers \cite{DBLP:journals/corr/abs-2107-01619} have found that leveraging user interactions would be a promising approach for reducing  color-breeding artifacts.
These methods require a certain amount of human effort, and the quality of results depends on the user's skills. Text-based methods usually adopt image captions \cite{DBLP:conf/naacl/ManjunathaIBD18} or palettes converted from the text \cite{DBLP:conf/eccv/BahngYCPWMC18} as means of intervention. However, the color represented by text is challenging to transfer to the image accurately.

The third class, denoted as exemplar-based method, is a trade-off between fully automatic and human intervention strategies. Compared to the above two classes, it adopts sample reference images to provide rich colors without requiring the user to do too much manual work. The key and difficulty of exemplar-based colorization is to establish an accurate correspondence between these two images. DEPN \cite{DBLP:journals/cgf/XiaoHZQWHH20} uses a pyramid structure to exploit multi-scale color information, but it only captures the global tones because no semantic correspondence is established. Some works \cite{DBLP:conf/cvpr/XuWFSZ20} regard exemplar-based colorization as a style transfer problem, but cannot guarantee the correctness of semantics because they also lack a correspondence. Deep Image Analogy \cite{DBLP:journals/tog/LiaoYYHK17} was used in \cite{DBLP:journals/tog/HeCLSY18} to make the target and reference luminance channels aligned to get a coarse chrominance map for further refinement. \cite{DBLP:conf/mm/LuYPZW20} uses features extracted from the luminance channel of the target and reference images to obtain dense correspondence. However, inaccuracies caused by using luminance channels to calculate correspondence and wrong matching problems introduced by dense correspondence will lead to unsatisfactory results.
A general attention based framework is proposed in \cite{DBLP:conf/mm/YinLZP21} to fuse colors from the database when the correspondence is not established. However, this method sometimes will mistakenly use the colors from the database when the selected two images are highly semantically related, resulting in the final results looking different from the reference image.

\section{Methods}
\subsection{Overview of the Proposed Method}
The task of exemplar-based colorization can be formulated as follows. Given a gray-scale image $I_g$, which only contains the luminance channel $l$, our goal is to predict the corresponding $a$ and $b$ color channels in the CIE Lab color space, according to the reference color image $I_r$. The main challenge is to build an appropriate correspondence between the gray-scale image and the color reference. In order to make full use of the color information in the reference image, we will utilize the reference image twice in a coarse-to-fine manner during the whole colorization process. The proposed framework, namely Semantic-Sparse Colorization Network (SSCN), consists of two auxiliary modules, which transfer global and local colors in the reference image, respectively.

\begin{figure}[t!]
  \centering
  \includegraphics[width=1\linewidth]{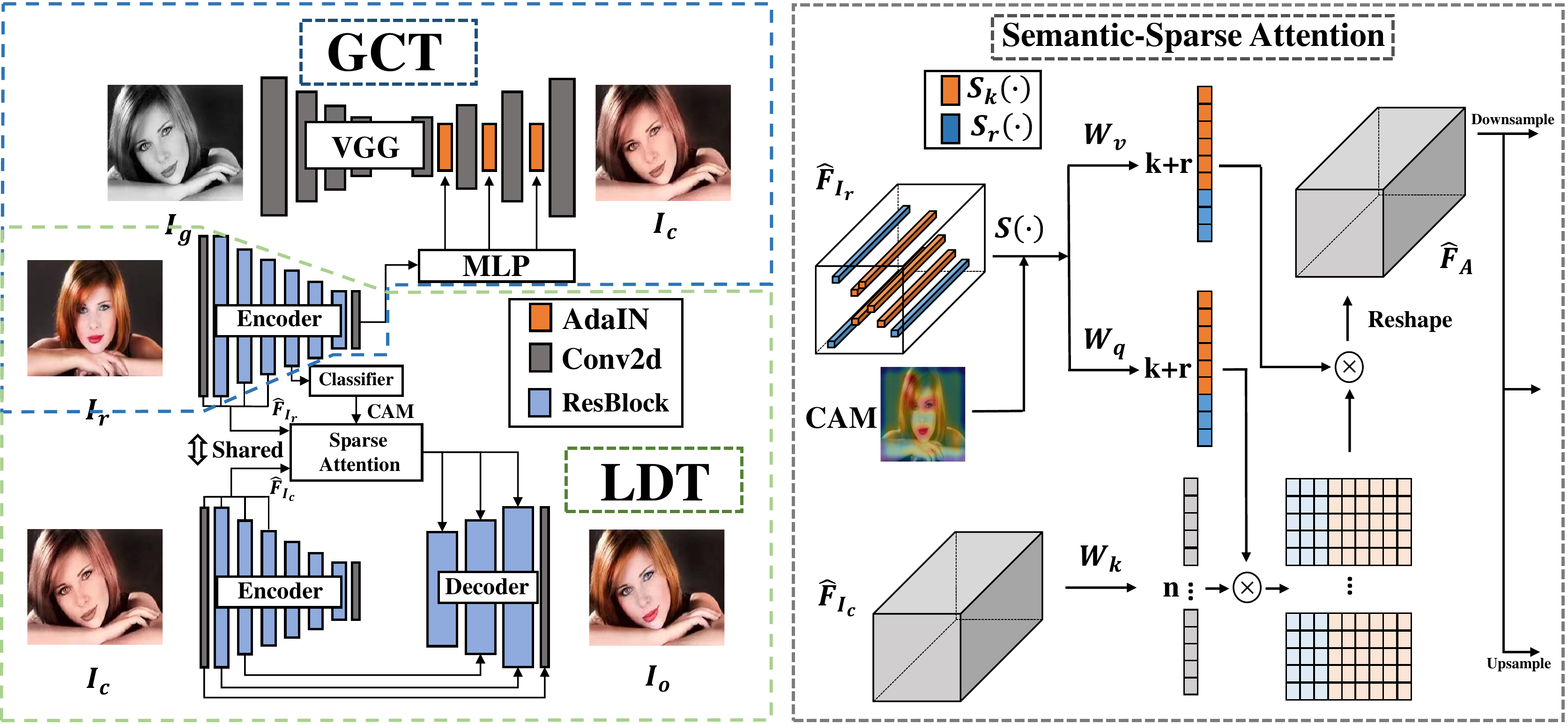}
  \caption{The illustration of the proposed two-stage image colorization framework. Our method uses a coarse-colorized image to build more accurate correspondence, which is completely different from previous works.
  The right part shows our proposed sparse attention mechanism in detail. With the help of the semantic information provided by CAM, the model can accurately use the critical parts of the reference image and reduce the complex computation caused by the attention mechanism.}
  \label{fig::model_archi}
\end{figure}

The overall pipeline of SSCN is illustrated in Figure \ref{fig::model_archi}. Specifically, taking the reference image $I_{r}$ as input, our model will first encode it into features $F_{I_{r}}$. These features will be used in both global and local coloring modules. In the coarse colorization stage, the Global Color Transfer (GCT) module will use $F_{I_{r}}$ to preliminarily color the gray-scale image $I_{g}$, and get a coarse-colorized result $I_{c}$, which has similar global tones as $I_{r}$. Then the coarse output $I_{c}$ will be further encoded into features $F_{I_{c}}$ with the same encoder as $ F_{I_{r}}$. In the fine colorization stage, the Local Details Transfer (LDT) module will use $F_{I_{r}}$ and $F_{I_{c}}$ to construct a correspondence that focuses on the semantically relevant regions of $I_{r}$. Note that these regions are sparsely selected according to their semantic levels. Based on the predicted mappings from LDT, the reference features $F_{I_{r}}$ are reorganized and fused with $F_{I_{c}}$ at different scales. Finally, the decoder takes the fused color features to produce the $a$ and $b$ channels of the input image $I_g$.


\subsection{Global Color Transfer}

We will first introduce the encoder of $I_{r}$, which is shared in both GCT and LDT modules. The encoder consists of six residual blocks. The last layer of $F_{I_{r}}$ is passed through an MLP to form the style vector, which will be used in the GCT module for global style transfer. In GCT, the gray-scale image $I_g$ will first be encoded into features $\{x_1,x_2,...,x_n\}$. Then, we perform coarse colorization in the feature space by changing feature statistics with AdaIN operation as:
\begin{equation}
  AdaIN(x_i,y)=y_{s,i}\frac{x_i-\mu(x_i)}{\sigma(x_i)}+y_{b,i}\ ,
  \label{equ:adain}
\end{equation}
where $\mu(x_i)$ and $\sigma(x_i)$ represent the $i^{th}$ feature map's mean and variance, respectively. $y_{s}$ and $y_{b}$ are the affine parameters of the style vector, which is obtained from $F_{I_{r}}$ via MLP transformation. Each feature map $x_i$ is normalized separately and then scaled/biased using the corresponding coefficients from $y(y_s,y_b)$. After affine transformation, each feature channel will have the activation for certain color information. These features can be inverted to the Lab space by a convolutional decoder. We finally get the coarse colorized result $I_{c}$ of the coarse colorization stage. In our implementation, the encoder uses sub-layers of the VGG19 \cite{DBLP:journals/corr/SimonyanZ14a}, and the decoder is symmetric structure. AdaIN are added after CNN layers of the decoder.

\subsection{Local Details Transfer}

The target of the LDT module is to build a more detailed and accurate correspondence between the coarse-colorized result $I_{c}$ and the reference image $I_{r}$. To begin with, we encode $I_{c}$ into the corresponding features $F_{I_{c}}$, with the same encoder as $F_{I_{r}}$. To find their correspondence, we extract features from the first four layers of $F_{I_{r}}$ and $F_{I_{c}}$, and resize them to the same spatial size of $1/4$ input image. Then these features are concatenated to form features $\hat{F}_{I_{r}}$ and $\hat{F}_{I_{c}}$, corresponding to the latent states of coarse and reference image, respectively. Their spatial size is both $d\times H/4\times W/4$, where $d$ is the number of feature maps. To facilitate computation, they are further flattened in the last two directions, and form features of size $d\times HW/16$. In this way, we segment the input image into $HW/16$ regions and represent each region with a $d$ dimensional vector. 

Based on the obtained features $\hat{F}_{I_{r}}$ and $\hat{F}_{I_{c}}$, the LDT module will calculate a correlation matrix $A$ via attention mechanism, whose element is computed by the scaled dot product \cite{DBLP:conf/nips/VaswaniSPUJGKP17} illustrated as Formula \ref{equ:attnmap}:

\begin{equation}
  \alpha_{ij}=\mathop{softmax}_{j}\left(\frac{(W_{q} f^{c}_i)\cdot (W_{k} f^r_j)}{\sqrt{d}} \right).
  \label{equ:attnmap}
\end{equation}

Here, $\alpha_{ij}$ represents the similarity between the $i$-th region of $\hat{F}_{I_{c}}$ and the $j$-th region of $\hat{F}_{I_{r}}$. $\hat{F}_{I_{c}}$ is used to retrieve relevant local details from $\hat{F}_{I_{r}}$. Then, we can re-weight the features $\hat{F}_{I_{r}}$ to obtain the attended feature $\hat{F}_{a}$ through a weighted sum operation as Formula \ref{equ:att}:
\begin{equation}
  f^a_i=\sum\limits_j \alpha_{ij}W_{v} f^r_j\ ,
  \label{equ:att}
\end{equation}
where $W_{q}$, $W_{k}$ and $W_{v}$ represent the linear transformation matrix into \textit{query}, \textit{key}, and \textit{value} vectors, respectively. The attended features $\hat{F}_{a}$ will be reshaped to the size of $d \times H/4 \times W/4$ and further resized into a suitable shape, fused with the features $F_{I_{c}}$ at different scales and fed into the U-Net \cite{DBLP:conf/miccai/RonnebergerFB15} decoder for the final detailed result of the fine colorization stage.

\begin{figure}[t!]
  \centering
  \includegraphics[width=1\linewidth]{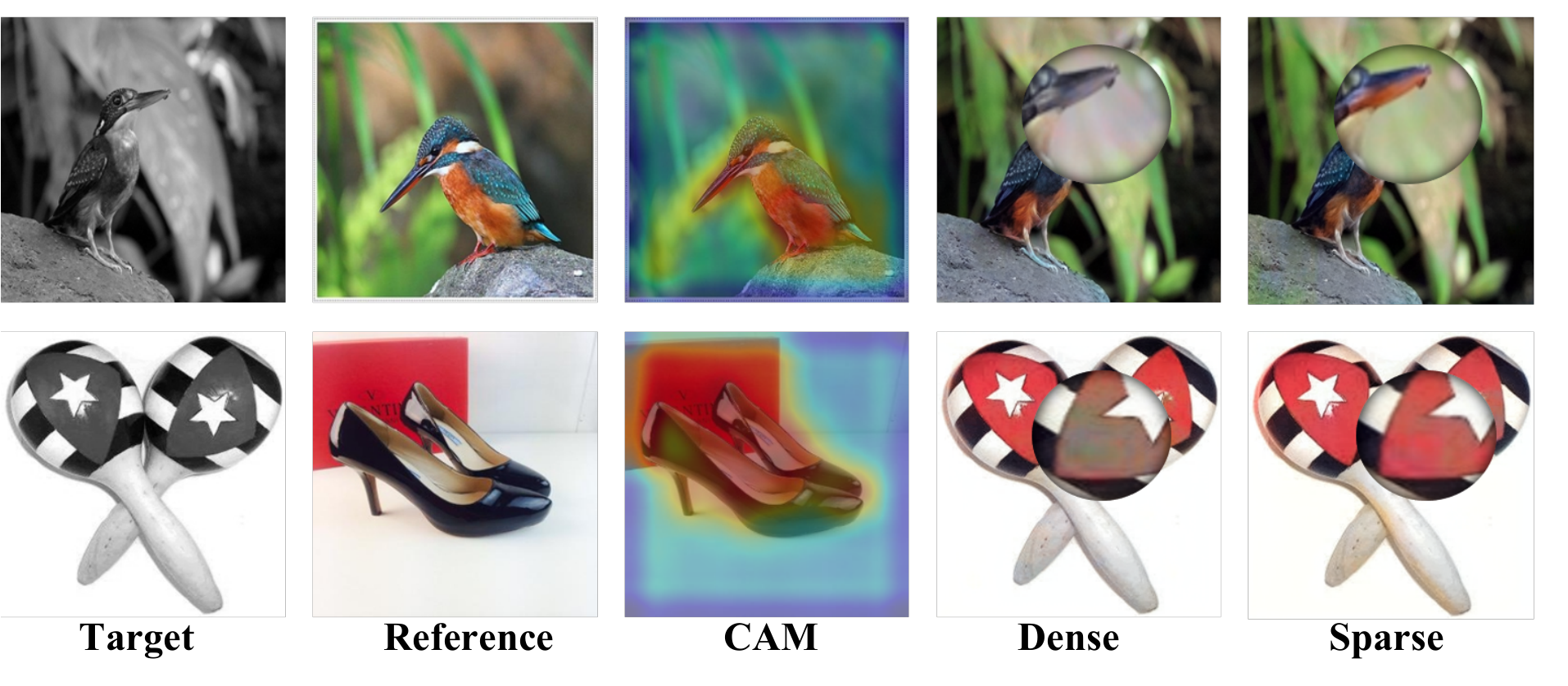}

  \caption{Comparison results of dense and sparse correspondence strategies.
  The output results will be disturbed by the re-weighting process using dense correspondence. Sparse attention focusing on semantically important areas can solve this problem.} 
  \label{fig::den_vs_spa}
\end{figure}

\textbf{Semantic-Sparse Correspondence.} In the above description, we use a standard attention mechanism to calculate the dense correspondence between coarse and reference images. We further propose a semantically sparse correspondence for better results with less computation cost. To be specific, the reference features $\hat{F}_{I_{r}}$ will go through a selection operation. First, the fifth layer of $F_{I_{r}}$ will be fed into a classifier and get a class activation map (CAM) \cite{DBLP:conf/cvpr/ZhouKLOT16}, which is used as the reference for selection. The CAM is flattened to $C =\{c_1,c_2,...,c_{HW/16}\}\in \mathbb{R}^{HW/16}$. The selection operation $S\left(\cdot\right)$ contains the top-$k$ selection $S_k\left(\cdot\right)$ and random selection $S_r\left(\cdot\right)$ implemented upon $C$. The $S_k\left(\cdot\right)$ selects the $k$ largest elements of $C$ and records their indexes $\mathbf{T}_k$. This encourages the attention mechanism to focus more on semantically significant areas and reduce the interference caused by insignificant parts. At the same time, the coloring of the background areas also needs reference. Thus $S_r\left(\cdot\right)$ randomly selects $r$ more indexes $\mathbf{T}_r$. Finally, we obtain $S\left(C\right) = \mathbf{T}_k\cup\mathbf{T}_r$ and the semantic-sparse features $\hat{F}_{I_{r}}\left[S\left(C\right)\right]$. To calculate the new correspondence map, we can simply replace the features $\hat{F}_{I_{r}}$ with $\hat{F}_{I_{r}}\left[S\left(C\right)\right]$ in Formula \ref{equ:attnmap},\ref{equ:att}. The other steps remain the same as above.

\subsection{Discussion}
\textbf{Dense Correspondence vs. Sparse Correspondence.}
Dense correspondence will be easily affected by irrelevant regions, especially when the reference is completely different from the gray-scale image. Even if the target region has low similarity with most reference regions, the re-weighting process will still disturb the final result.
In contrast, sparse correspondence can overcome this difficulty by focusing only on semantically important regions, which can reduce the interference of other regions. Moreover, the computational complexity goes from $\mathcal{O}\left((HW)^2\right)$ to $\mathcal{O}\left((k+r)HW\right)$, while $(k+r)$ is generally 8 to 16 times smaller than $HW$.
The comparison results of these two strategies are shown in Figure \ref{fig::den_vs_spa}.
It can be observed that some details are more accurately colorized after reducing the interference.

\begin{figure}[t!]
  \centering
  \includegraphics[width=1\linewidth]{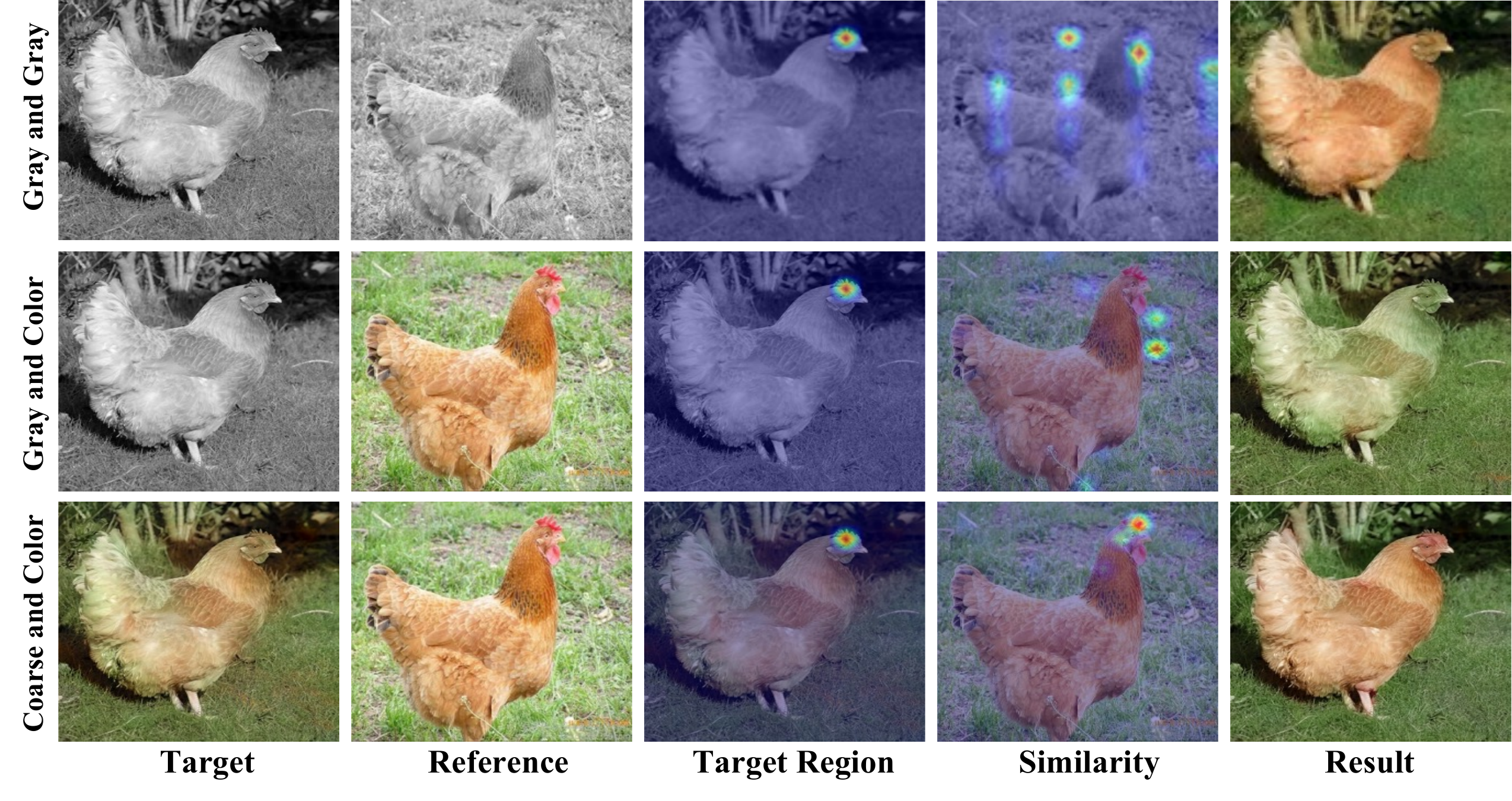}

  \caption{Comparison results of using three different data types to build the correspondence. The coarse-colorized result we proposed to use can establish more accurate correspondence than the other two common types.}
  \label{fig::attention_com}
\end{figure}

\textbf{Coarse-colorized vs. Gray-scale.}
In this work, we propose to use a coarse-colorized image to build the correspondence with the reference, which is completely different from previous works \cite{DBLP:conf/mm/LuYPZW20,DBLP:conf/mm/YinLZP21,DBLP:conf/cvpr/ZhangHLSYBC19,DBLP:journals/corr/abs-2201-04364}. The coarse result is already consistent with the reference's global color style, thus can produce more dedicated correspondence than directly using the gray-scale image. Moreover, the correspondence between color images is more accurate than that between gray-scale images (luminance channels). To verify this comment, we build a correlation matrix for three data types with the same operations. Figure \ref{fig::attention_com} shows the comparison results of the similarity between one target region and all reference regions. It is clear that the chicken comb is correctly matched between two color images, even with different colors. 

\subsection{Objective Functions}
\subsubsection{\textbf{Smooth-L1 Loss.}}
To avoid simply using the average scheme for solving the ambiguity colorization problem, a widely used loss function Smooth-L1 loss is adopted in image colorization tasks. This loss is added to the results of both two stages in our architecture as $L_{stage1}$ and $L_{stage2}$. The following Formula \ref{equ:sl1} can calculate the Smooth-L1 loss between $T_{ab}$ and $\hat{T}_{ab}$:

\begin{small}
\begin{equation}
 L_{stage1,2}(T_{ab},\hat{T}_{ab})=    \left\{
\begin{aligned}
&\frac{1}{2}(T_{ab}-\hat{T}_{ab})^2 \quad for \left |T_{ab}-\hat{T}_{ab}\right | \leq \delta \\
& \delta \left |T_{ab}-\hat{T}_{ab}\right | -\frac{1}{2} \delta^2 \quad otherwise.
\end{aligned}
\right.
\label{equ:sl1}
\end{equation}
\end{small}

\subsubsection{\textbf{Classification Loss.}}
There is a classification loss $L_{cls}$ in the classifier to get a CAM as a reference for $S\left(\cdot\right)$. This loss can also improve the encoder's ability of extracting color features.
When $F_{I_{r}}$ is fed into the classifier, its label vector is predicted. $L_{cls}$ is defined as the cross-entropy between the classification vector $\hat{y}$ and its ground truth one-hot label.

\subsubsection{\textbf{Color Histogram Loss.}}
To transfer the color distribution of the reference image to the target image accurately, we also add a histogram loss to the final output as Formula \ref{equ:his}. Similar to the previous work \cite{DBLP:conf/eccv/ZhangIE16}, we treat the problem as multinomial classification. We quantify $\hat{T}_{ab}$ output space into bins with $grid size=10$ and keep the in-gamut $Q = 313$. The mapping to predicted color distribution $\hat{Z}\in [0,1]^{H\times W\times Q}$ is also learned with the decoder. 
The $L_{his}$ is defined as a cross-entropy loss for every pixel to measure the distance between predicted distribution $\hat{Z}$ and ground truth $Z$, and sum over all pixels.
\begin{equation}
 L_{his}(\hat{Z},Z)=-\sum\limits_{h,w} \sum\limits_{q}Z_{h,w,q}\log (\hat{Z}_{h,w,q})\ .
\label{equ:his}
\end{equation}
\subsubsection{\textbf{TV Regularization.}}
To encourage spatial smoothness in the output result $\hat{T}_{ab}$, we follow previous work \cite{DBLP:conf/eccv/JohnsonAF16} and apply the total variation regularization $L_{TV}(\hat{T}_{ab})$ to the output of the fine colorization stage. 

In summary, the overall loss function for the entire network is defined as: 
\begin{equation}
\begin{aligned}
    L_{total}&=\lambda_{stage1}L_{stage1}+\lambda_{stage2}L_{stage2} +\lambda_{TV}L_{TV}+\lambda_{cls}L_{cls}+\lambda_{his}L_{his} \ ,
\end{aligned}
\label{equ:total}
\end{equation}
where $\lambda_{stage1}$, $\lambda_{stage2}$,  $\lambda_{TV}$, $\lambda_{cls}$ and $\lambda_{his}$ are hyperparameters to constrain different loss terms.

\section{Experiments}
\subsection{Implementation Details}
We use ImageNet's \cite{DBLP:conf/cvpr/DengDSLL009} total training set to train the entire network with $5$ epochs and set mini-batch size as $8$. During training, the input image will be resized to $256\times256$. We use Adam \cite{DBLP:journals/corr/KingmaB14} for optimization with $\beta_1=0.9, \beta_2 = 0.999$. The learning rate is set to $0.0001$. 
We set the coefficients for each loss function as follows: $\lambda_{stage1} = 100$, $\lambda_{stage2}= 100$, $\lambda_{cls} = 0.1$, $\lambda_{TV} = 10$, and $\lambda_{his} = 1$.
For the $S\left(\cdot\right)$, both $k$ and $r$ are set to $256$.

For the exemplar-based colorization method, it is impossible to find enough source-reference pairs to train the network. 
We adopt a scheme similar to \cite{DBLP:conf/cvpr/LeeKLKCC20}. The reference is generated from the original image by geometric distortion, 
which can provide complete color information for the target image.
The geometric distortion is realized by thin plate splines (TPS) transformation.
The distortion is randomly applied to each image. 
In the training process, we apply violent transformation to some images to simulate semantically unrelated reference images. 


\subsection{Comparison with Previous Methods}
\subsubsection{\textbf{Visual Comparison.}}

\begin{figure*}[t!]
  \centering
  \includegraphics[width=\linewidth]{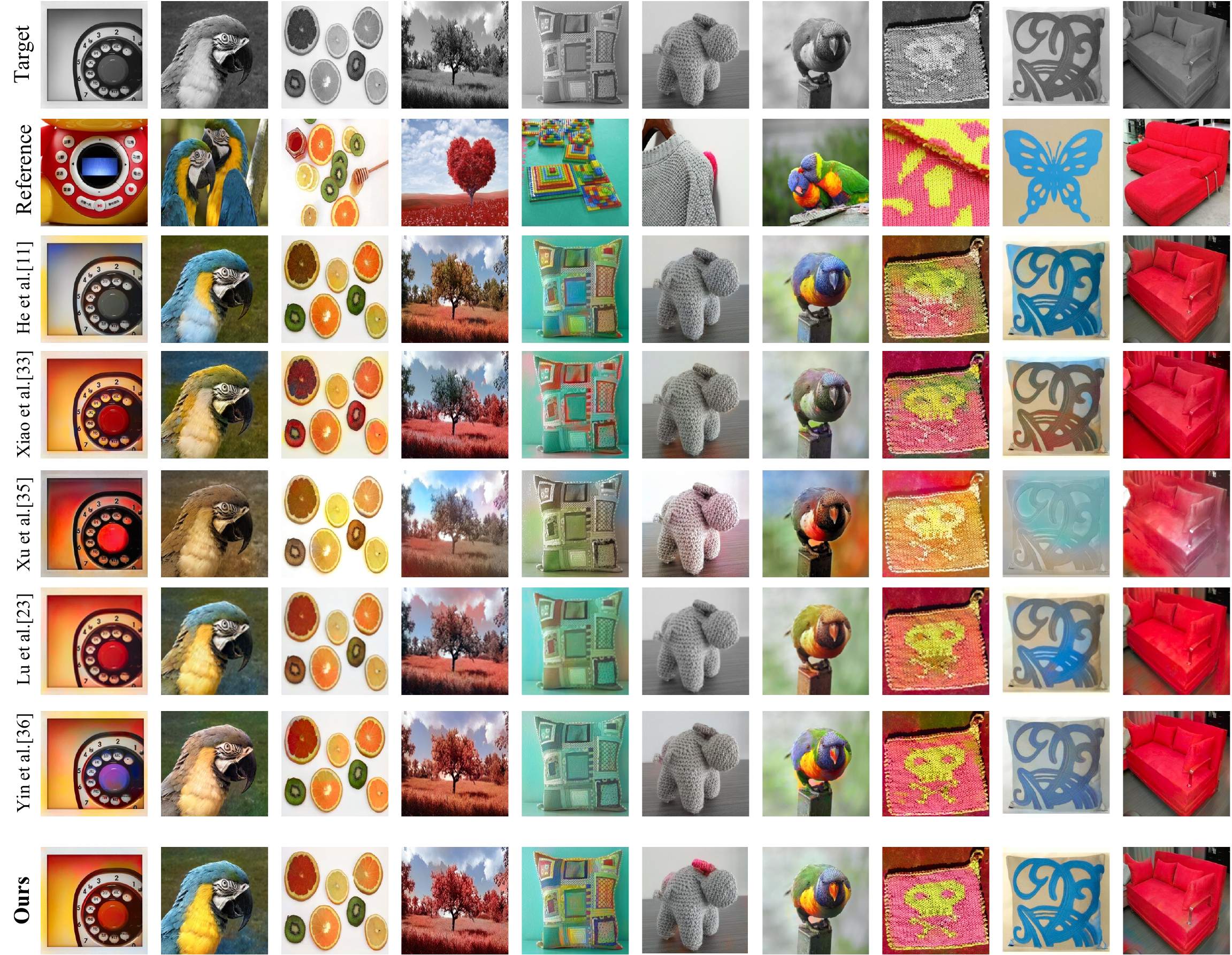}
  \vspace{-0.5cm}
  \caption{Qualitative comparison of colorizing results with previous methods. The target image, reference image, and each method's colorized images are displayed from top to bottom. The proposed method outperforms other models and achieves state-of-the-art performance.}
  \label{fig:cmp1}
  \vspace{-0.3cm}
\end{figure*}

We compare the results of our method with previous exemplar-based colorization approaches \cite{DBLP:journals/cgf/XiaoHZQWHH20,DBLP:conf/cvpr/XuWFSZ20,DBLP:journals/tog/HeCLSY18,DBLP:conf/mm/LuYPZW20,DBLP:conf/mm/YinLZP21}. We run all $6$ models on $230$ pairs of images collected from ImageNet validation set and show several representative results. All comparison results are obtained by public available codes. We show the qualitative comparison in Figure \ref{fig:cmp1}. See our supplementary materials for more results.

The 4th column of Figure \ref{fig:cmp1} shows the results of colorizing objects with unusual or artistic colors. Compared with method \cite{DBLP:journals/tog/HeCLSY18} constrained by the perceptual loss, the proposed method can appropriately colorize the target image according to the user's requirement. Since \cite{DBLP:conf/mm/LuYPZW20} tends to make the color histograms of the two images consistent, resulting in the wrong spatial distribution of colors.

In the 6th column, when there are large regions with less semantics in the image, our method can pay more attention to the semantically relevant areas, e.g., the the pink area, while other methods fail to colorize the object or simply get a smooth result. In the 2nd column, the parrots in two input images are highly semantically related, while \cite{DBLP:conf/mm/YinLZP21} uses the colors in the database, resulting in an unsatisfactory final result.

When the reference image is semantically unrelated to the target image (shown as 1st column in Figure \ref{fig:cmp1}), due to the dependence on prior color knowledge, \cite{DBLP:journals/tog/HeCLSY18} will ignore the colors from the reference image. Histogram-based methods \cite{DBLP:journals/cgf/XiaoHZQWHH20} can get plausible results by transferring global tones, whereas our method can yield better results.
For some images with many details,  \cite{DBLP:conf/mm/LuYPZW20} cannot properly colorize these details due to the inappropriate correspondence constructed with two gray-scale images, while the proposed method allows the target image to be colored correctly, e.g., 5th and 7th columns in Figure \ref{fig:cmp1}.

\begin{figure}[t!]
  \centering
  \includegraphics[width=1\linewidth]{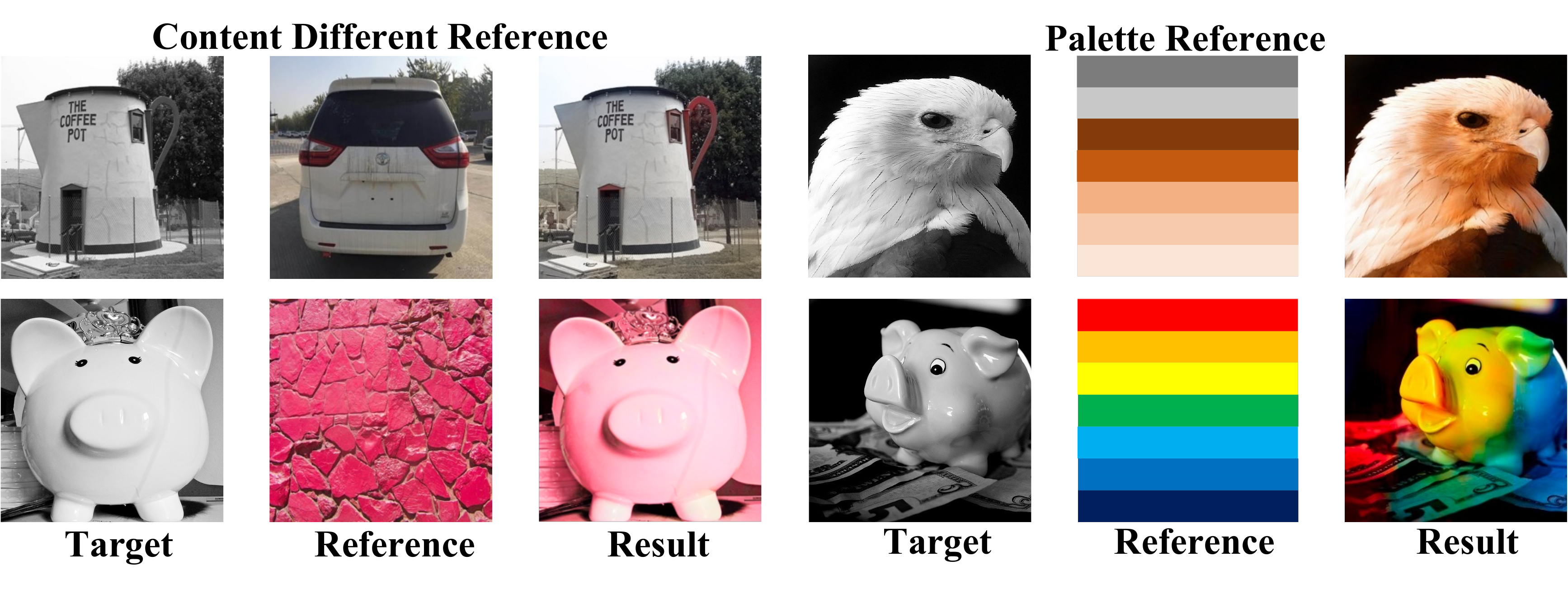}

  \caption{Colorization results of using content different references and palettes. Visually satisfactory results can also be obtained using these two types of references.}
  \label{fig::palette}
\end{figure}

These experimental results show that the proposed method can transfer color information for different image pairs accurately and effectively. We also show some results of using content different references and palettes in Figure \ref{fig::palette}. Even when the semantics of the reference image are irrelevant or have no semantics, our method can also get satisfactory results.

\begin{figure}[t!]
  \centering
  \includegraphics[width=1\linewidth]{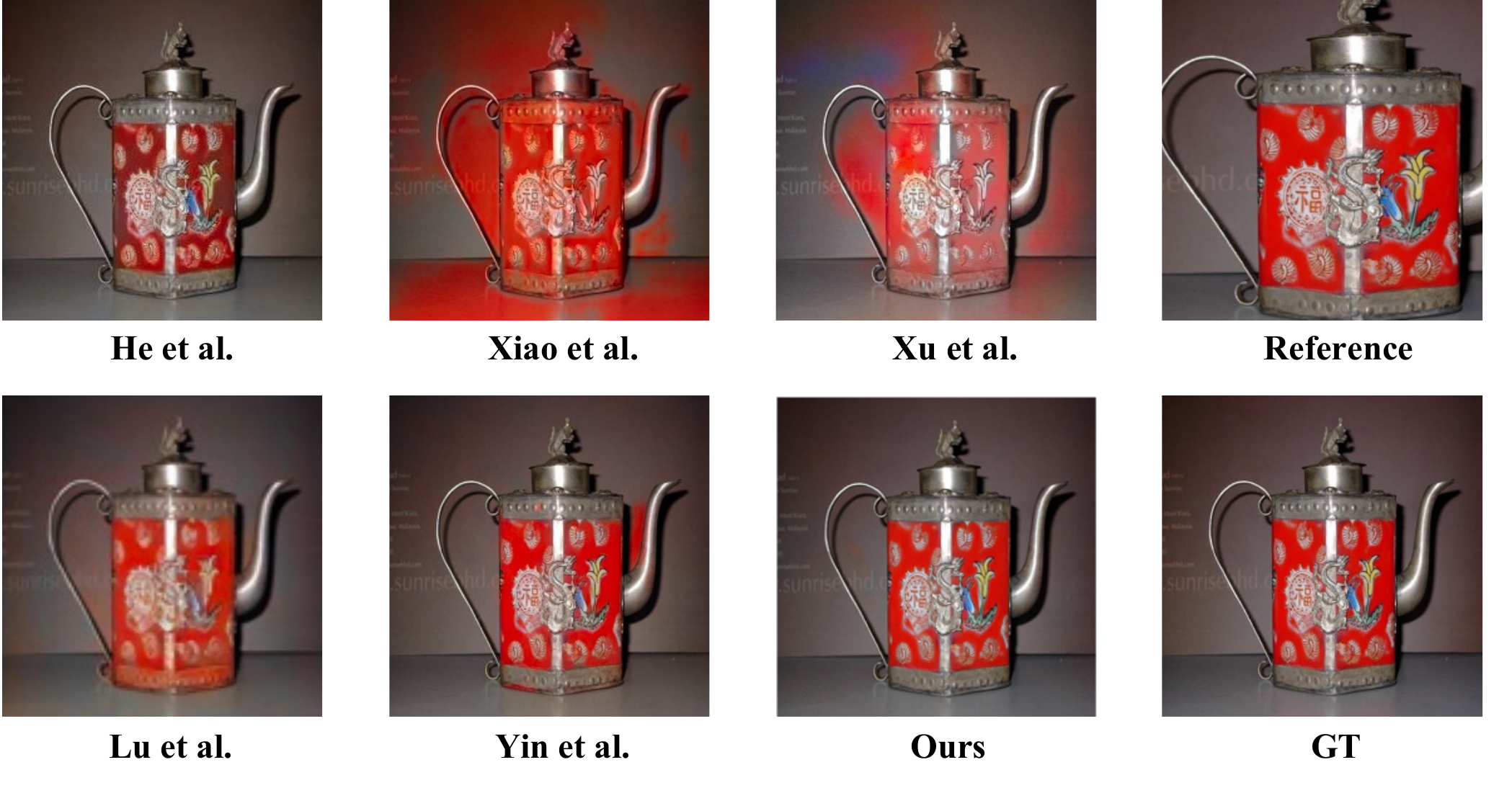}

  \caption{Comparison results of using random cropping reference image to colorize the target image. The results obtained by other methods are not satisfactory even when using such a suitable reference.
  }
  \label{fig::psnr_vis_com}
\end{figure}

\subsubsection{\textbf{Self-Augmentation PSNR/SSIM.}}
Unlike automatic colorization, in exemplar-based colorization setting, when given a target-reference pair, there is no ground truth that has both the target's shape and the reference's color. 
The histogram intersection similarity (HIS) used in previous work \cite{DBLP:conf/mm/LuYPZW20,DBLP:conf/mm/YinLZP21} is not a suitable index. Mismatches may also occur in the spatial color distribution of the result with high histogram similarity with the reference image.
In order to make a quantitative evaluation of the colorization results, similar to the training process, we use the augmentation of a color image as the reference to colorize its luminance channel, so that the original color image can be used as ground truth. With ground truth available for comparison, some existing evaluation metrics, such as peak signal-to-noise ratio (PSNR) and structural similarity (SSIM), can be used for evaluation.

We select $5000$ images from the validation set of ImageNet to do three different data augmentation, including TPS, random rotation (RR), and random cropping (RC) as references to get different results. The quantitative comparisons of three different augmentation are reported in Table \ref{Tab.result}. Figure \ref{fig::psnr_vis_com} shows an example of using a RC reference and comparing the results with other methods.
We will release this test dataset for future comparison.

\begin{table}[t!]
    \centering
    \scriptsize  
        \caption{Quantitative comparisons of self-augmentation PSNR/SSIM. A higher value indicates a better preference, while the proposed method outperforms other models.}
    \resizebox{\linewidth}{!}{%

    \begin{tabular}{c|c|c|c|c}
     \toprule
    \hline

    \textbf{Methods}      & \makecell[c]{TPS} & \makecell[c]{RR}   & \makecell[c]{RC}   & \makecell[c]{\textbf{Mean}}
    
    \\\hline 

    He et al.(2018)\cite{DBLP:journals/tog/HeCLSY18}  & 28.51/0.902 & 28.67/0.903 & 27.57/0.898 & 28.25/0.901  \\
    
    Xiao et al.(2020)\cite{DBLP:journals/cgf/XiaoHZQWHH20}     & 25.17/0.912 & 25.30/0.913 & 24.98/0.910 & 25.15/0.911  \\
    
    Xu et al.(2020)\cite{DBLP:conf/cvpr/XuWFSZ20}                 & 22.46/0.873 & 21.65/0.846 & 21.55/0.862  & 21.88/0.860  \\
    
    Lu et al.(2020)\cite{DBLP:conf/mm/LuYPZW20}               & 27.93/0.913 & 29.80/0.931 & 27.12/0.907 & 28.28/0.917\\ 
    
    Yin et al.(2021)\cite{DBLP:conf/mm/YinLZP21}               & 31.87/0.948 & 34.24/0.952 & 29.85/0.939 & 31.98/0.946

    \\ 
    \textbf{Ours}              & \textbf{36.32/0.969} & \textbf{35.49/0.966} & \textbf{32.39/0.958} & \textbf{34.73/0.964} 
     \\ 
    
    \hline
    \bottomrule
    \end{tabular}
   
    }

    \label{Tab.result}
\end{table}

\subsubsection{\textbf{User Evaluation.}}
We conduct user evaluation to verify the proposed method's effectiveness subjectively. In this part, we randomly select $50$ groups from the above results. Semantically dependent pairs and semantically unrelated pairs are distributed in half. Eventually, all $6\times50$ color images are distributed anonymously and randomly to $30$ college participants.

For fairness, the images with the same reference are shown simultaneously in a random order. All participants were asked to observe the images for no more than $5$ seconds and choose the image that better matches the reference. As shown in Figure \ref{fig::pie}, we show the percentage of votes for each method in the form of pie chart. It shows that images of our method are mostly preferred.



\vspace{0.5cm}

\makeatletter\def\@captype{figure}\makeatother
\begin{minipage}{.45\textwidth}
	
	\includegraphics[width=0.94\linewidth]{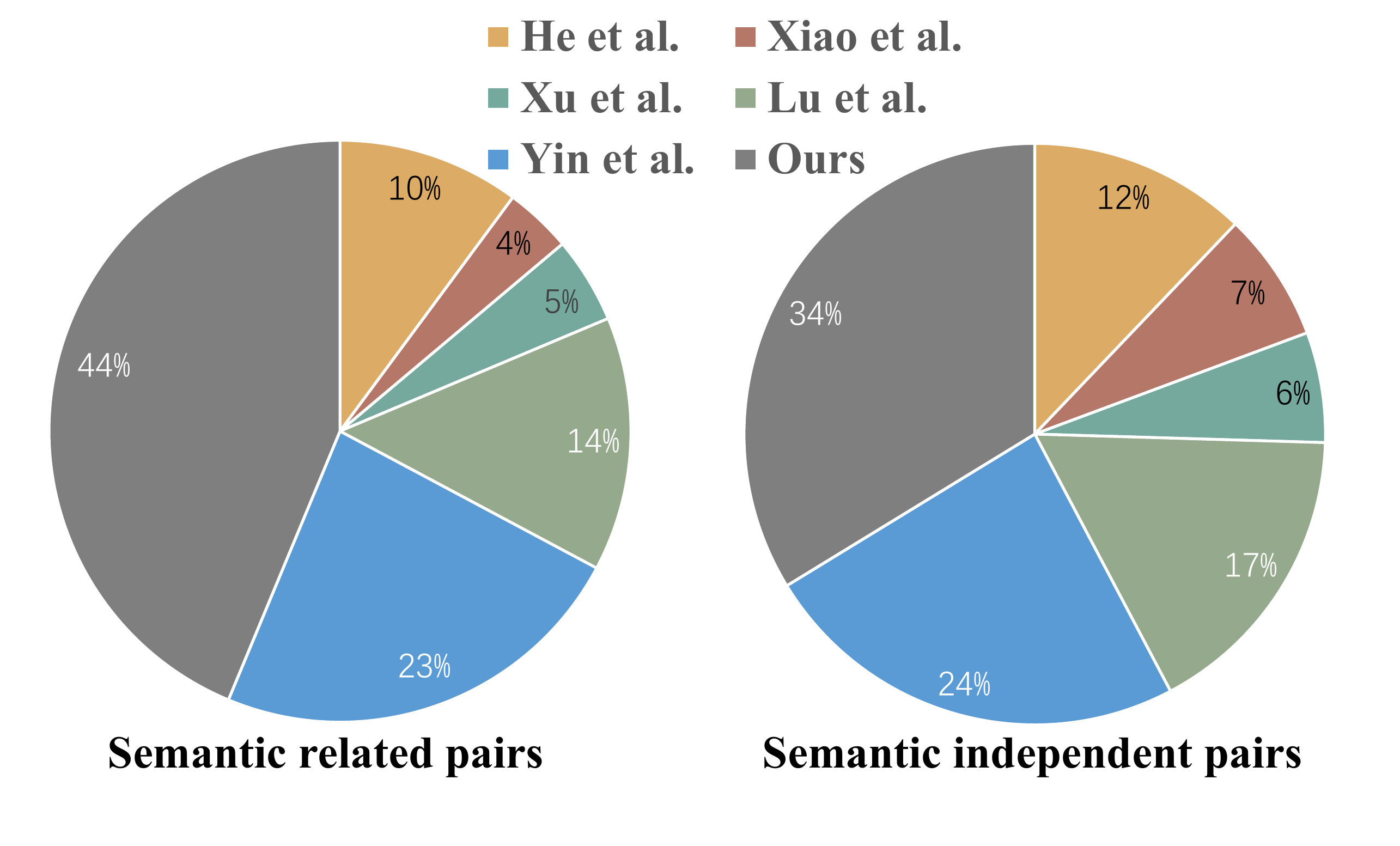} 
	\caption{The users' preferences for six different methods. Under two different image pairs, our results have been the most selected by users. }
	\label{fig::pie} 
\end{minipage}
\quad
\makeatletter\def\@captype{figure}\makeatother
\begin{minipage}{.45\textwidth}
	
\includegraphics[width=0.94\linewidth]{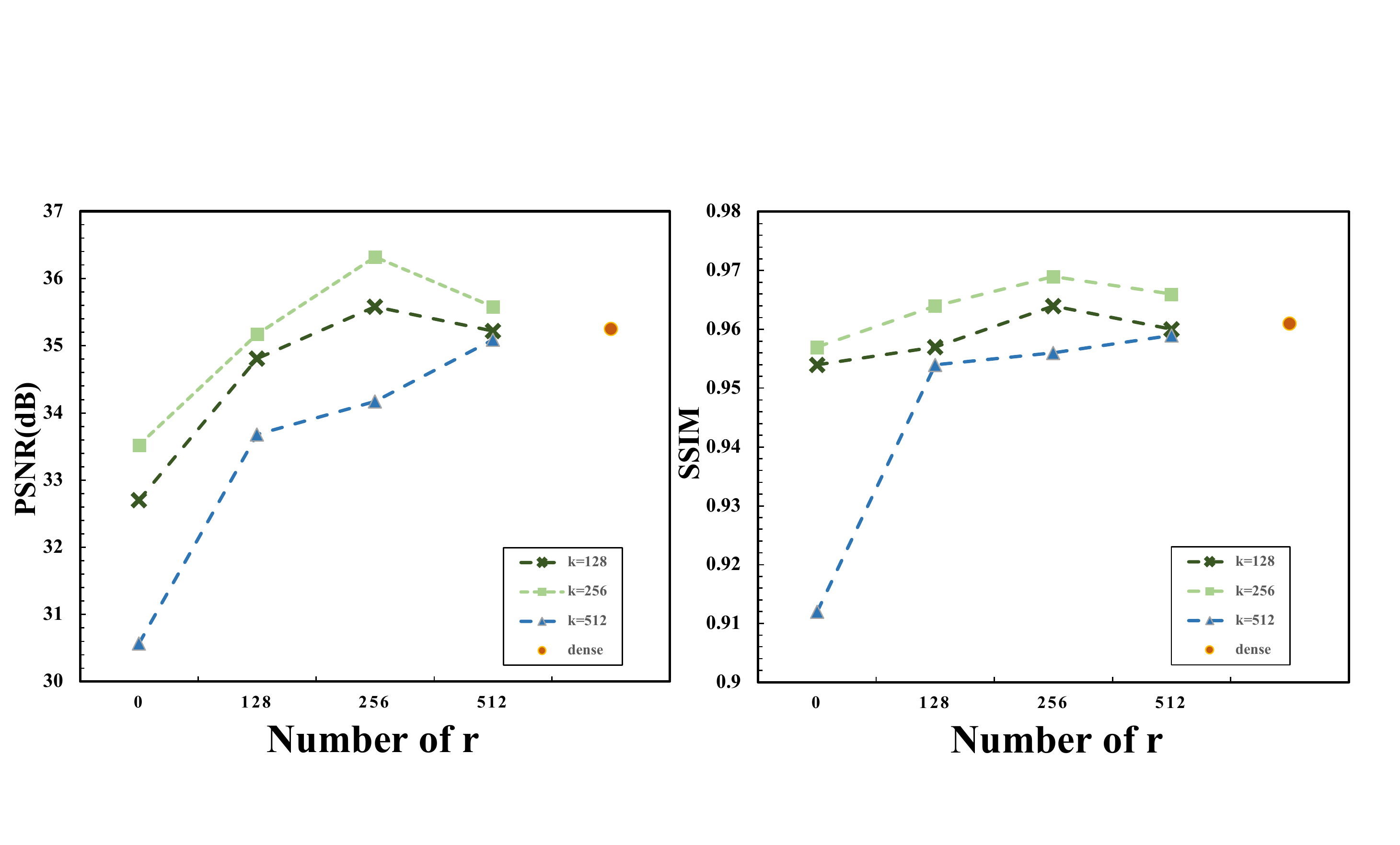} 
	\caption{ The picture shows how the numerical changes of $k$ and $r$ affect the final result. Larger or smaller $k$ and $r$  will reduce the quality.}
	\label{fig::line}
\end{minipage}

\section{Ablation Studies}

\subsubsection{Ablation study of \textbf{$S_k\left(\cdot\right)$ and $S_r\left(\cdot\right)$}.}
The use of sparse correspondence will lead to the question: how to select an appropriate number of regions in the process?
Then we further study the effect of $k$ and $r$, and use TPS reference to evaluate results quantitatively as described above. When the resolution of the reference image is $256\times 256$, there are $4096$ features available for selection. We increase $k$ and $r$  gradually from $128$ and $0$, respectively. The comparison results are shown in Figure \ref{fig::line}. Without random selection, the value of PNSR/SSIM will be much lower because some areas of the background are incorrectly colored. Increasing $r$ gradually can improve the results, but increasing $r$ further will cause the result deteriorate again. In addition, it can be seen from the comparison of the three broken lines that a larger or smaller $k$ will reduce the quality.
    
    

\subsubsection{\textbf{Ablation study of two-stage architecture.}}

\begin{figure}[t!]
  \centering
  \includegraphics[width=1\linewidth]{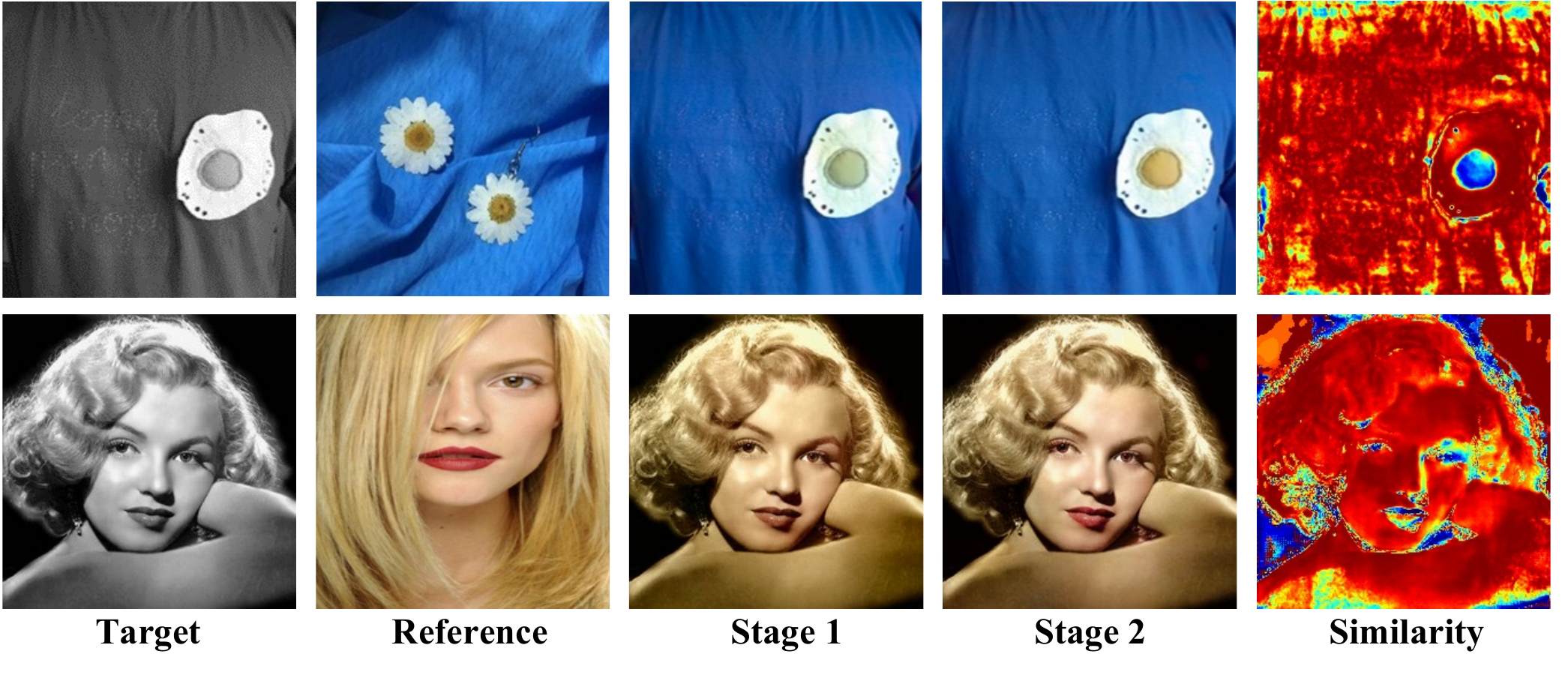}

  \caption{Ablation study on the relationship between the results of the two stages. The bluer the part in the heat map, the less similar the features. The main differences are concentrated in some semantic details.}
  \label{fig::2stage}
 
\end{figure}

To illustrate the importance of the two-stage structure in our model, we conduct ablation study on $k=256, r=256$ version. 
First, we evaluate the first stage results with PSNR and SSIM values of $30.02$ and $0.937$. There is a huge gap between them and the final results, thus illustrating the importance of LDT. 
To further validate the importance of preliminary coloring, we remove GCT from the whole architecture for comparison. 
Instead, we use another network with a similar structure to the encoder of $I_{r}$ but with one channel input to extract the features of gray-scale image and calculate the correspondence in the same way. Due to the lack of information in the gray-scale image, the PSNR and SSIM values will decrease by $3.30$ and $0.014$.
We also analyze the relationship between the results of the two stages in the encoder feature space. The similarity of features is shown in the form of heat map in Figure \ref{fig::2stage}. We can see that the differences between the two are mainly concentrated in some semantic details, which are completed in the second stage.

\subsubsection{Ablation study of Loss Functions.}In order to verify that the classifier does not only provide a CAM but also help the encoder extract color features, we ablate the classification loss on the dense version. 
After this loss is removed, the corresponding PSNR and SSIM values are $33.73$ and $0.952$, while the PSNR and SSIM values of the dense version are $35.25$ and $0.961$.
In addition, we also ablate color histogram loss of the best version ($k=256, r=256$) to analyze its effect. The PSNR and SSIM values will decrease by $1.28$ and $0.009$.
Removing either of these losses will reduce the model's performance, especially in the $L_{cls}$.

\section{Conclusions}
This paper proposes a colorization framework named Semantic-Sparse Colorization Network (SSCN) to colorize the target image in a coarse-to-fine manner.
Specifically, an image transfer network is adopted in the coarse colorization stage to obtain a preliminary colorized result. In the fine colorization stage, semantically related areas of the reference image
will be selected to to color the details of the target image. Thus, SSCN can adequately transfer a reference image's global color and local details onto a gray-scale image. It provides a way to obtain different levels of color information from the reference image hierarchically and accurately. Extensive experiments show that the proposed method outperforms previous state-of-the-art approaches by a large margin.

\myparagraph{Acknowledgments.} 
This work was supported by SZSTC Grant No.JCYJ201908
09172201639 and WDZC20200820200655001, Shenzhen Key Laboratory ZDSYS2
0210623092001004.
%
%
\bibliographystyle{splncs04}
\bibliography{egbib}
\end{document}